\documentclass{article}

\usepackage[main, final]{neurips_2026}

\usepackage[utf8]{inputenc}
\usepackage[T1]{fontenc}

\usepackage{amsmath}
\usepackage{amssymb}
\usepackage{amsfonts}

\usepackage{microtype}
\usepackage{xcolor}
\usepackage{graphicx}
\usepackage{booktabs}
\usepackage{nicefrac}

\usepackage{multirow}
\usepackage{tabularx}

\usepackage{hyperref}
\usepackage{url}
\usepackage{wrapfig}
\usepackage{algorithm}
\usepackage{algpseudocode}
\usepackage[most]{tcolorbox}
\usepackage{listings}
\newtcblisting{promptbox}{
    colback=gray!5,
    colframe=black,
    coltitle=black,
    fonttitle=\bfseries,
    title=Prompt,
    arc=2mm,
    boxrule=0.6pt,
    breakable,
    listing only,
    listing options={
        basicstyle=\ttfamily\footnotesize,
        breaklines=true,
        columns=fullflexible
    }
}

\newcommand{\method}{COMPASS}

\title{COMPASS: Composable Policy-Amortized Structured Search for LLM-Based Optimization Modeling}

\author{%
  Nguyen Le Minh Hoang\textsuperscript{1} \quad
  Van Duc Cuong\textsuperscript{12}%
  \thanks{Co-first author.} \quad
  Bui Trong Duc\textsuperscript{1}%
  \thanks{Corresponding author.} \quad
  Huynh Thi Thanh Binh\textsuperscript{1} \\
  \textsuperscript{1}Hanoi University of Science and Technology, Hanoi, Vietnam\\
  \textsuperscript{2}VinMotion, Hanoi, Vietnam\\
  \texttt{hoang.nlm2417125@sis.hust.edu.vn, cuong.vd220021@sis.hust.edu.vn}\\
\texttt{ducbt@soict.hust.edu.vn, binhht@soict.hust.edu.vn}
}

\begin{document}
\maketitle
\begin{abstract}
Large language models offer a promising technique for translating natural-language optimization problems into solver code, but their reliability remains constrained by the quality of template retrieval. While recent structured methods improve over flat prompting through hierarchical taxonomy search, their traversal rules remain fixed or LLM-driven at inference time, leaving retrieval unable to improve from past successes and failures. This motivates treating template retrieval as a policy that can be learned from modeling experience, rather than as a static inference-time rule. We propose COMPASS, a policy-amortized framework for retrieving optimization templates in LLM-based solver modeling, which represents the template library as a directed acyclic graph and formulates retrieval as a sequential decision process over template nodes. A deep Q-network learns a reusable traversal policy from template-level judgments and execution feedback, reducing reliance on fixed heuristic traversal through feedback-driven graph navigation. To support extensible retrieval, COMPASS embeds problem descriptions and template nodes in a shared semantic space, allowing newly inserted templates to be scored from their text descriptions without redefining the action space. Experiments across multiple LLM backbones and optimization-modeling benchmarks demonstrate that COMPASS consistently improves over heuristic structured retrieval, with the largest gains on complex and mixed-category problems.
\end{abstract}

\section{Introduction}
\label{sec:introduction}

Large language models (LLMs) are increasingly used to translate natural-language
optimization problems into mathematical formulations and executable solver code.
This setting, often referred to as autoformulation, has the potential to lower
the expertise barrier for building operations-research models
\citep{astorga2025autoformulation,NEURIPS2025_aeaf9b45,NEURIPS2025_903475a4}.
However, optimization problems rarely follow a single canonical pattern. A
single description may combine routing, assignment, scheduling, capacity, and
resource-allocation structures, making correct template selection a central
bottleneck for downstream code generation.

Recent methods improve reliability by injecting structured modeling knowledge
into the prompting process. Autoformulation approaches decompose problems into
variables, objectives, and constraints, while structured retrieval methods organize modeling knowledge into hierarchical taxonomies
\citep{astorga2025autoformulation,NEURIPS2025_aeaf9b45,NEURIPS2025_903475a4}.
Yet these retrieval strategies are typically fixed at inference time. They can
guide an LLM toward relevant modeling patterns, but they do not learn from
which templates previously led to valid formulations, feasible solver runs, or
execution failures. In contrast, search-based methods can adaptively explore
alternative formulations, but doing so separately for each instance incurs
substantial test-time cost.

This paper is motivated by the view that optimization-template retrieval is not
only a semantic matching problem, but a sequential decision process over
structured modeling knowledge. Early retrieval choices determine which modeling
families and executable templates remain reachable, so a good retriever should
both exploit the organization of the template library and improve from past
modeling feedback. We propose \method, a policy-amortized framework that
represents the template library as a directed acyclic graph (DAG) and learns a
traversal policy from LLM-based structural judgments and solver-execution
signals. By scoring problem-template compatibility in a shared semantic space,
\method\ can evaluate newly inserted templates from their textual metadata
without changing the policy architecture, while beam traversal enables retrieval
of complementary templates for mixed-structure problems.

Our main contributions are:
\begin{itemize}
    \item We recast optimization-template retrieval as a graph-structured
    sequential decision problem over a template DAG, where internal nodes
    represent modeling families and leaves represent executable templates. The
    DAG captures shared substructures among optimization models and allows a
    template to belong to multiple related modeling paths, making the retrieval structure more flexible than a strict tree taxonomy.
    \item We develop a policy-amortized retrieval algorithm that learns reusable
    traversal decisions from LLM-based structural judgment and solver-execution
    feedback. The method uses semantic node scoring to evaluate candidate
    templates from textual metadata, enabling new templates to be inserted
    without redefining the policy output space.
    \item We empirically validate \method\ across multiple optimization-modeling
    benchmarks and LLM backbones, demonstrating consistent improvements over direct
    prompting, heuristic structured retrieval, and test-time search baselines,
    with the largest gains on structurally complex problems.
\end{itemize}

\section{Related Work}
\label{sec:related}

LLM-based optimization modeling has recently advanced through autoformulation,
structured retrieval, and feedback-guided search \citep{van2026single}. AutoFormulator explores the
formulation space with MCTS and symbolic pruning
\citep{astorga2025autoformulation}, SolverLLM uses test-time MCTS with dynamic
expansion, prompt backpropagation, and uncertainty-aware feedback
\citep{NEURIPS2025_903475a4}, and OptiTree retrieves modeling thoughts from a
hierarchical OR taxonomy \citep{NEURIPS2025_aeaf9b45}. Related systems extend
this direction to mathematical modeling and OR reasoning, including MATHMO for
sequential decision-making under uncertainty \citep{liu2026mathmo}, StepORLM
for reinforcement learning and generative process supervision
\citep{zhou2026steporlm}, and operator-theoretic autoformulation for
Bellman-equation operator graphs \citep{baillet2026operator}. Beyond
autoformulation, LLMs have also been used as optimizers or heuristic designers:
OPRO performs optimization through iterative prompting
\citep{yang2024large}, ReEvo combines evolutionary search with LLM reflection
\citep{NEURIPS2024_4ced59d4}, and HiFo-Prompt improves long-term knowledge reuse
in automated heuristic design \citep{chentongchen2026hifoprompt}. More broadly,
LLM reasoning systems use search, decomposition, tool feedback, and memory to
improve inference-time decisions, as in Tree of Thoughts
\citep{NEURIPS2023_271db992}, self-consistency \citep{wang2023selfconsistency},
Auto-CoT \citep{zhang2023automatic}, least-to-most prompting
\citep{zhou2023leasttomost}, ReAct \citep{yao2023react}, CRITIC
\citep{gou2024critic}, Self-Refine \citep{NEURIPS2023_91edff07}, and Reflexion
\citep{NEURIPS2023_1b44b878}. Retrieval-augmented and tool-using systems such
as Self-RAG \citep{asai2024selfrag}, DSPy \citep{khattab2024dspy}, ToolLLM
\citep{qin2024toolllm}, DRAFT \citep{qu2025from}, and ToolEVO
\citep{chen2025learning} further show the value of grounding LLM behavior in
external evidence and execution feedback. In contrast to methods that rely on
fixed traversal rules, costly per-instance search, or prompt/tool optimization,
\method\ learns a reusable semantic value function over a persistent template
DAG, allowing structured optimization-template retrieval to improve from solver
and judge feedback while remaining efficient at inference time. More related
work is presented in Appendix~\ref{app:related}.

\section{Methodology}
\label{sec:methodology}

\begin{figure}
    \centering
    \includegraphics[width=\linewidth]{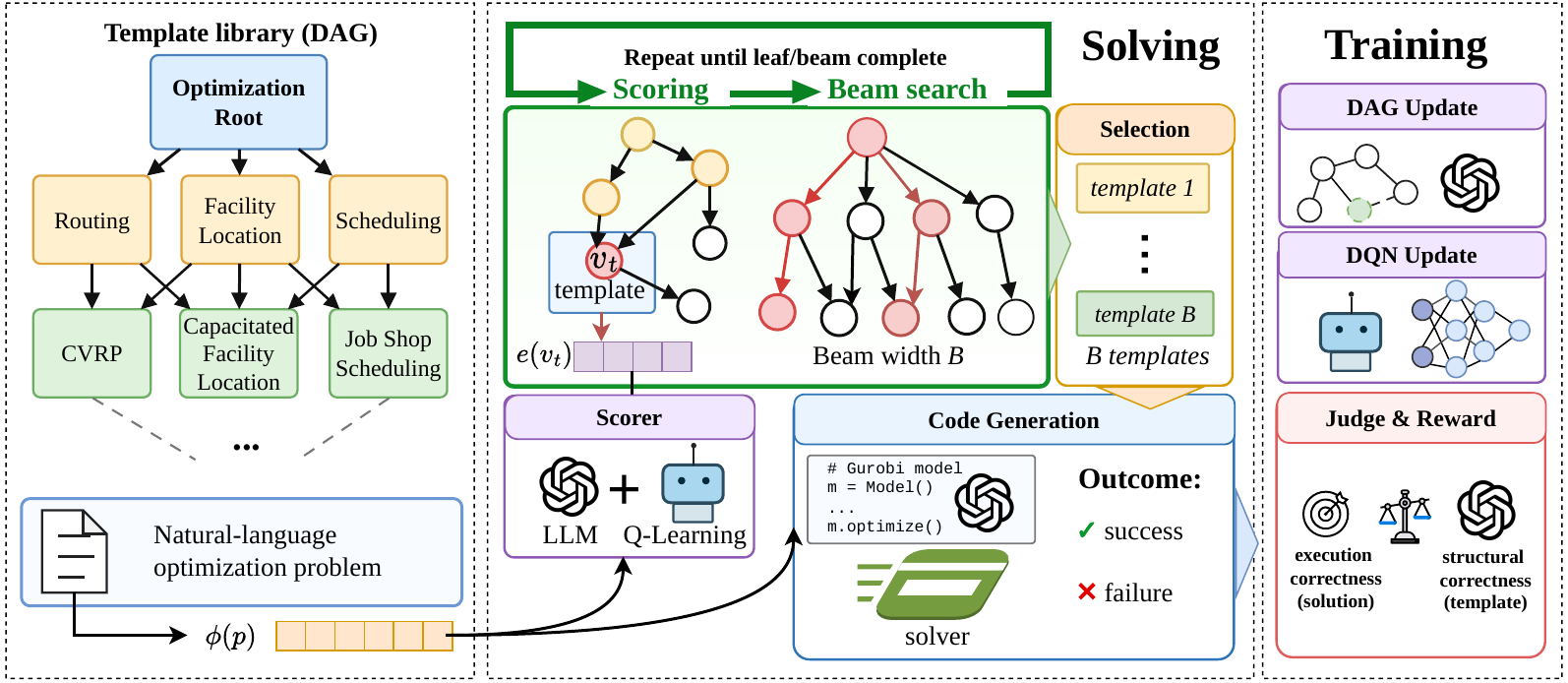}
    \caption{Overview of \method. The template library is represented as a DAG, and retrieval is performed through a learned traversal policy guided by semantic priors and execution feedback.}
    \label{fig:compass-overview}
\end{figure}

We propose \method, a retrieval-augmented framework for translating
natural-language optimization problems into executable solver programs. The
core difficulty lies in formulation selection: before solver code can be
generated, the system must identify suitable modeling abstractions, variables,
objectives, and constraints. Since early formulation choices determine which
templates and constraint structures remain plausible, template retrieval cannot
be treated as independent semantic matching. Instead, retrieval should model
the dependencies among successive formulation decisions. Figure~\ref{fig:compass-overview} illustrates this formulation-selection process and shows how \method\ connects graph-based retrieval, template composition, solver-code generation, and feedback-driven policy improvement. Algorithmic details of the inference and training procedures are provided in
Appendix~\ref{app:algorithms}.

\method\ represents the formulation library as a DAG rather
than a flat set of templates or a strict taxonomy. Each node denotes a modeling
state with associated textual metadata, and each edge specifies a valid
refinement from a broader abstraction to a more specialized formulation. This
representation captures the fact that optimization models often share
intermediate structures: for example, routing, assignment, and capacity
constraints may co-occur across otherwise distinct problem classes. Allowing
multiple incoming paths to a node enables such shared modeling patterns to be
reused without duplicating templates.

Given a problem description $p$, the system computes a semantic embedding
$x=\phi(p)$ and evaluates candidate graph nodes during traversal. The learned
value function estimates the expected usefulness of selecting a node with
respect to downstream formulation quality, rather than relying only on local
textual similarity. A language-model semantic prior provides an additional
compatibility signal when execution feedback is sparse. The final retrieval
procedure combines these scores within beam search to identify one or more
templates that can jointly support solver-code generation.

Training uses supervision from structural judgments and solver execution. After
a retrieved template is instantiated and evaluated, the resulting feedback is
assigned to the visited trajectory, so both terminal templates and intermediate
modeling decisions receive credit. This mechanism converts retrieval into an
experience-driven policy over the graph: previous successes and failures shape
future traversal, while inference remains substantially cheaper than repeated
per-instance search.

\subsection{Structured Retrieval over a Formulation DAG}

We organize the template library as a directed acyclic graph
$\mathcal{G}=(\mathcal{V},\mathcal{E})$, where each node represents a reusable
formulation abstraction and each edge encodes a valid refinement relation
between modeling concepts. A retrieval trajectory therefore corresponds to the
progressive specialization of a modeling hypothesis: starting from broad
formulation families, such as routing, scheduling, or resource allocation, the
policy moves toward more specific abstractions and eventually reaches executable
template nodes.

At each retrieval step, the traversal state is defined as $s=(x,\phi(v))$, where $v\in\mathcal{V}$ is the current graph node, and $\phi(v)$ is the semantic embedding. The admissible action set is
restricted to the children of the current node, $A(s)=\mathcal{C}(v)$, where $\mathcal{C}(v)=\{u\in\mathcal{V}\mid (v,u)\in\mathcal{E}\}$. This
constrained action space preserves structural consistency among successive
formulation decisions and reduces the effective search space relative to flat
selection over all templates in the library.

The DAG representation plays a key role in modeling the non-mutually exclusive
structure of optimization knowledge. In contrast to a strict tree taxonomy,
where each abstraction has a single parent, a DAG permits multiple refinement
contexts for the same formulation node. This is important because modeling
concepts such as capacity, precedence, assignment, and flow conservation often
recur across different optimization families. The resulting structure supports
parameter sharing across related formulations, avoids redundant template
duplication, and defines a structured decision space suitable for learning
long-horizon retrieval policies.

\subsection{Semantic Value Function}

The traversal policy should remain valid when the formulation library evolves.
To this end, \method\ represents both problem descriptions and formulation
nodes in a shared semantic space, rather than binding actions to fixed template
identifiers. Let $\tau(\cdot)$ denote the tokenization map, $F_{\eta}$ a Transformer encoder,
and $P$ a pooling operator. For a text sequence $z$, we define
$h_z=P(F_{\eta}(\tau(z)))$ and
$\phi(z)=h_z/(\|h_z\|_2+\epsilon)$, where $\epsilon>0$ is a small constant for
numerical stability. The problem embedding and node embedding are given by
$x=\phi(p)$ and $e_v=\phi(m_v)$, respectively, where $m_v$ denotes the textual
metadata associated with node $v$. For each candidate node $v$, its semantic representation is projected into the
policy space as $r_v=\sigma(W_p e_v+b_p)$, where
$\sigma(a)=\max(a,0)$ is applied elementwise. The semantic value function is
then defined as $Q_{\theta}(x,e_v)=q_{\theta}([x\parallel r_v])$, where
$q_{\theta}$ is a scalar-valued neural function and $[\cdot\parallel\cdot]$
denotes vector concatenation.

This parameterization makes the traversal policy depend on semantic
compatibility between the input problem and candidate formulation node, rather
than on memorized template indices. Consequently, related formulations can share
statistical strength, and newly inserted nodes can be embedded and evaluated
directly from their metadata without modifying the output dimension of the value
function.

\subsection{Inference: Value-Guided Traversal with LLM Regularization}

At inference time, \method\ selects child nodes using a score that combines the
learned value estimate with a language-model compatibility prior. For a
candidate child $c\in\mathcal{C}(v)$, we define
$\psi(c\mid p)=Q_{\theta}(x,e_c)+\lambda \ell(c,p)$, where
$\ell(c,p)\in[0,1]$ denotes the semantic compatibility assigned by the language
model and $\lambda\ge0$ controls the strength of this prior.

The language-model prior is used as a regularizer rather than as the main
retrieval policy. Since $Q_{\theta}$ is trained from a finite set of execution
trajectories, its estimates may be less reliable for rare, underexplored, or
newly inserted nodes. The prior $\ell(c,p)$ provides a zero-shot semantic signal
over optimization concepts and formulation patterns, while $Q_{\theta}$ encodes
task-specific experience accumulated from previous solver executions. Their
combination improves traversal robustness under sparse feedback without
reverting to costly language-model-driven search.

To handle problems with multiple interacting modeling structures, \method\
performs beam traversal with beam size $k$. The procedure maintains the top-$k$
partial paths according to $\psi$ and returns a set of candidate leaves
$\mathcal{L}(p)$. The beam width is selected on validation instances to balance
coverage and computational cost. This is especially useful for hybrid
optimization problems whose complete formulation may require complementary
templates from different refinement paths.

The retrieved leaves are converted into templates and composed into a final
formulation $t^{\star}=\mu(\{t_v:v\in\mathcal{L}(p)\})$, where $\mu$ denotes the
template composition operator. The resulting formulation is then instantiated
into executable solver code.

\subsection{Learning from Execution Feedback}

We model DAG traversal as a deterministic finite-horizon contextual decision
process. Each episode starts from a problem embedding $x=\phi(p)$, follows a
trajectory over formulation nodes, and terminates with template instantiation,
solver-code generation, and execution. For a terminal node $v$, the reward is
defined as $R(p,v)=\alpha R_{\mathrm{judge}}(p,v)+(1-\alpha)R_{\mathrm{exec}}(p,v)$,
where $R_{\mathrm{exec}}$ measures solver-level correctness and
$R_{\mathrm{judge}}$ assesses structural compatibility between the problem and
the retrieved formulation. This decomposition mitigates the sparsity of pure
execution feedback by providing an additional semantic signal when generated
programs fail syntactically, numerically, or due to incomplete constraints.

The optimal node-value function satisfies
$Q^\star(x,e_v)=\mathbb{E}[R(p,v)]$ for $v\in\mathcal{T}$ and
$Q^\star(x,e_v)=\gamma\max_{c\in\mathcal{C}(v)}Q^\star(x,e_c)$ otherwise,
where $\mathcal{T}$ denotes the set of terminal nodes and $\gamma$ mildly
penalizes unnecessarily long refinement paths. Bellman propagation assigns
credit from terminal execution outcomes to earlier modeling abstractions,
allowing the policy to learn not only which templates succeed, but also which
intermediate decisions tend to lead to valid formulations.

We train the value network with temporal-difference learning. For transition
$i$, the target is
$y_i=r_i+(1-d_i)\gamma\max_{c\in\mathcal{C}(v_i)}Q_{\theta^-}(x_i,e_c)$,
where $d_i$ indicates termination and $\theta^-$ denotes the target network.
The training objective is
$\mathcal{L}(\theta)=\mathbb{E}_{i\sim\mathcal{B}}[w_i\ell_\delta(Q_\theta(x_i,e_{v_i})-y_i)]$,
where $\ell_\delta(a)=\frac{1}{2}a^2$ if $|a|\le\delta$ and
$\ell_\delta(a)=\delta(|a|-\frac{1}{2}\delta)$ otherwise. This smooth
$L_1$ objective improves stability under noisy execution and judge feedback.

To improve sample efficiency, we use prioritized experience replay. Each
transition is sampled with probability
$P(i)=p_i^{\alpha_{\mathrm{PER}}}/\sum_j p_j^{\alpha_{\mathrm{PER}}}$, where
$p_i=|\Delta_i|+\varepsilon_{\mathrm{PER}}$ and $\Delta_i=Q_\theta(x_i,e_{v_i})-y_i$.
The corresponding importance weight is
$w_i=(1/(N P(i)))^\beta/\max_j(1/(N P(j)))^\beta$. Prioritizing high-error
transitions focuses updates on underexplored or poorly estimated regions of the
traversal space, which is particularly important under sparse terminal rewards.
Hyperparameters such as $\lambda$, $k$, $\alpha_{\mathrm{PER}}$, and $\beta$ are
selected on validation instances.

\subsection{Adaptive Template Expansion}

Execution failures can also indicate limitations of the formulation library,
rather than only errors in the traversal policy. To address this case,
\method\ allows the formulation graph to expand when repeated failures suggest
that existing templates do not sufficiently cover a recurring modeling pattern. Expansion is enabled only after the retrieval policy has reached a minimum
level of stability. Specifically, we use the gate
$g_t=\mathbb{1}(t\ge\tau_{\min})\mathbb{1}(\varepsilon_t\le\tau_\varepsilon)$, where $t$ denotes the training step, $\varepsilon_t$ is the exploration rate,
and $g_t\in\{0,1\}$ indicates whether expansion is permitted. This condition
prevents early exploration noise and unstable value estimates from producing
redundant or low-quality templates. Once the gate is active, persistent
execution failures are treated as evidence of missing formulation abstractions.

When expansion is triggered, the language model abstracts reusable formulation
patterns from failed instances and reference solutions, and proposes new
template nodes with textual metadata. These nodes are inserted into
$\mathcal{G}$ and embedded through the same semantic encoder used for existing
nodes, allowing the value function to score them without changing its output
space. Subsequent feedback then determines whether the newly added nodes become
useful retrieval targets. DAG expansion is performed only on training instances.

Thus, \method\ adapts along two complementary axes: the traversal policy
improves through execution-driven value learning, while the formulation graph
evolves when the current template set lacks sufficient expressive coverage.
This converts template retrieval from a fixed-library procedure into an
adaptive structured search process.

\section{Experiments}
\label{sec:experiments}

\subsection{Experimental Setting}

We evaluate whether \method\ improves the reliability, compositionality, and transferability of LLM-based optimization modeling through learned structured retrieval.

\paragraph{Benchmarks and backbones.}
We conduct experiments on six benchmarks: NL4Opt (245 problems)\citep{nl4opt}, MAMO-Easy (652)\citep{mamo}, MAMO-Complex (211)\citep{mamo}, LPWP (288)\citep{coe}, IndustryOR (100)\citep{coe}, and OptMATH \citep{optmath} (166) \citep{optmath}. These benchmarks cover linear programming, mixed-integer formulations, mathematical word problems, and industrial optimization scenarios, providing a diverse testbed for heterogeneous and compositional modeling. We use DeepSeek-V3 and GPT-5-mini as primary code-generation backbones, and evaluate transfer to Qwen-32B, LLaMA-8B, and DeepSeek-R1. This setup separates the effect of retrieval from the choice of downstream.

\paragraph{Baselines.}
We compare \method\ with direct prompting, OptiTree \citep{NEURIPS2025_aeaf9b45}, SolverLLM \citep{NEURIPS2025_903475a4}, and AutoFormulator \citep{astorga2025autoformulation}. We also include DAG-Greedy, which uses the same template library and generator as \method\ but replaces the learned value function with the LLM-derived prior. This baseline isolates the contribution of policy learning over the shared DAG structure.

\paragraph{Metrics.}
We report exact accuracy, which requires both successful execution and correct solutions, and code execution rate (CER), which measures runtime validity independently of solution correctness.

\paragraph{Research questions.}
Our evaluation addresses the following questions:
\begin{itemize}
    \item \textbf{RQ1: Effectiveness.} Does \method\ improve accuracy and
    robustness over existing LLM-based optimization modeling approaches?

    \item \textbf{RQ2: Structured retrieval.} How much does DAG-based structured
    retrieval contribute compared to flat or unstructured retrieval?

    \item \textbf{RQ3: Ablation.} How does each component of \method\ affect
    overall performance?

    \item \textbf{RQ4: Sensitivity.} How sensitive is \method\ to key design
    choices such as beam width, regularization strength, and replay
    configuration?
\end{itemize}

\subsection{Experimental Results}

\begin{table}[ht!]
\centering
\caption{Comparison of modeling accuracy (\%) between \method\ and existing
baselines. Best results are shown in \textbf{bold}; second-best results are
\underline{underlined}.}
\label{tab:deepseek-main}
\resizebox{\textwidth}{!}{
\begin{tabular}{@{}clcccccc@{}}
\toprule
\multirow{2}{*}{\textbf{Model}} & \multirow{2}{*}{\textbf{Method}} & \multirow{2}{*}{\textbf{NL4Opt}} & \textbf{MAMO} & \textbf{MAMO} & \multirow{2}{*}{\textbf{LPWP}} & \multirow{2}{*}{\textbf{IndustryOR}} & \multirow{2}{*}{\textbf{OptMATH}} \\
& & & \textbf{EasyLP} & \textbf{ComplexLP} & & & \\
\midrule
\multicolumn{8}{c}{\textit{Reasoning LLMs}}\\
\midrule
Reasoning & DeepSeek-R1 & 91.8 & 80.8 & 57.3 & 69.4 & 72.0 & \underline{50.0} \\
LLMs      & OpenAI-o1 & 92.7 & 85.4 & \underline{57.8} & 70.5 & 74.0 & 47.6 \\
\midrule
\multicolumn{8}{c}{\textit{Prompt-based Methods}}\\
\midrule
\multirow{5}{*}{DeepSeek-V3}
& Direct  & 89.4 & 78.8 & 53.1 & 67.7 & 54.0 & 40.9 \\
& SolverLLM & 92.2 & 80.8 & 55.5 & 69.1 & 62.0 & 43.4 \\
& AutoForm. & 91.4 & 79.7 & 54.1 & 68.4 & 57.0 & 41.6 \\
& OptiTree & 93.4 & 82.8 & 56.4 & 71.2 & 65.0 & 49.4 \\
& \method  & \underline{95.1} & 85.9 & \textbf{61.6} & \textbf{72.6} & 66.0 & \textbf{50.6} \\
\midrule
\multirow{5}{*}{GPT-5-mini}
& Direct  & 89.7 & 82.2 & 49.7 & 64.6 & 67.0 & 36.7 \\
& SolverLLM & 92.2 & 85.7 & 52.1 & 67.4 & 71.0 & 40.4 \\
& AutoForm. & 90.6 & 84.2 & 50.7 & 66.3 & 69.0 & 39.8 \\
& OptiTree & 93.4 & \underline{87.4} & 53.5 & 68.9 & \underline{75.0} & 43.3 \\
& \method  & \textbf{95.9} & \textbf{88.2} & \underline{57.8} & \underline{72.2} & \textbf{78.0} & 48.0 \\
\bottomrule
\end{tabular}%
}
\end{table}

 \paragraph{Main results.} Table~\ref{tab:deepseek-main} evaluates whether policy-guided DAG retrieval
improves LLM-based optimization modeling over direct prompting and existing
retrieval baselines. Across most datasets, \method\ achieves the highest
accuracy, with the strongest gains on structurally complex benchmarks such as
MAMO-Complex, IndustryOR, and OptMATH. With DeepSeek-V3, \method\ improves over
the strongest retrieval baseline by $5.2\%$ on MAMO-Complex and $3.1\%$ on
MAMO-Easy. With GPT-5-mini, the gains are larger on several datasets, including
$4.3\%$ on MAMO-Complex, $3.3\%$ on LPWP, $3.0\%$ on IndustryOR, and $4.7\%$ on
OptMATH.

The results demonstrate a clear pattern: the advantage of \method\ becomes more
pronounced as formulation complexity increases. Direct prompting and flat
retrieval are more susceptible to early modeling errors, which can propagate
into invalid constraints or incorrect solver code. In contrast, \method\
performs progressive refinement over a compositional DAG, enabling intermediate
modeling hypotheses to be assessed before final template selection. The
consistent gains over reasoning-oriented LLMs further suggest that scaling
reasoning ability alone is insufficient when the retrieved formulation scaffold
is inadequate. These findings substantiate our central premise that optimization modeling is
best treated as a structured retrieval problem. By learning a reusable traversal
policy from solver and judge feedback, \method\ enhances formulation
consistency and final solver accuracy, particularly on problems requiring
compositional modeling.

\begin{table}[ht!]
\centering
\small
\caption{Accuracy (Acc) and code execution rate (CER, \%) of \method\ across
different LLM backbones.}
\label{tab:llm-generalization}
\begin{tabular}{lcccc}
\toprule
& \textbf{NL4Opt} & \textbf{MAMO-Easy} & \textbf{MAMO-Complex} & \textbf{IndustryOR} \\
\textbf{Models} & Acc / CER & Acc / CER & Acc / CER & Acc / CER \\
\midrule
Qwen-32B (COMPASS)  &80.8 / 100.0 & 76.2 / 100.0 & 29.9 / 100.0 & 48.0 / 100.0\\
Qwen-32B (Direct)   &72.7 / 100.0 & 65.3 / 100.0 & 20.4 / 100.0 & 32.0 / 98.0 \\
LLaMA-8B (COMPASS)  &55.9 / 100.0 & 66.9 / 100.0 & 24.2 / 100.0 & 15.0 / 98.0\\
LLaMA-8B (Direct)   &51.4 / 100.0 & 16.1 / 98.6 & 10.4 / 97.2 & 10.0 / 96.0 \\
DeepSeek-R1 (COMPASS) &93.8 / 100.0 & 90.4 / 100.0 & 66.8 / 100.0 & 75.0 / 100.0 \\
DeepSeek-R1 (Direct)  &91.8 / 100.0 & 80.8 / 100.0 & 57.3 / 100.0 & 72.0 / 100.0 \\

\bottomrule
\end{tabular}
\end{table}

\paragraph{Generalization across LLM backbones.}
We further evaluate whether \method\ transfers across generator families and
model scales by pairing the same retrieval policy with Qwen-32B, LLaMA-8B, and
DeepSeek-R1. Table~\ref{tab:llm-generalization} indicates that \method\
consistently improves over direct prompting across all evaluated backbones, with
particularly large gains for weaker generators, such as $+50.8\%$ on MAMO-Easy
with LLaMA-8B and $+16.0\%$ on IndustryOR with Qwen-32B.

These results suggest two conclusions. First, the learned retrieval policy
transfers across model architectures, indicating that the benefit of structured
formulation retrieval is largely backbone-agnostic. Second, smaller generators
benefit more substantially from DAG-guided retrieval, suggesting that \method\
serves as an external modeling scaffold that reduces the formulation burden on
the downstream LLM. In addition, \method\ achieves near-perfect code execution
rates in most settings, indicating that structured retrieval improves not only
solution accuracy but also the syntactic and structural reliability of generated
solver programs.

To assess whether these gains are merely due to additional inference
computation, we compare \method\ against repeated direct prompting under matched
inference budgets. Increasing the sampling or prompting budget alone does not
recover the improvements obtained by \method, indicating that the gains arise
from policy-guided formulation selection rather than brute-force inference
scaling.

\paragraph{Learned traversal versus heuristic retrieval.}

\begin{table}[ht!]
\centering
\small
\caption{Accuracy (\%) comparison between \method\ and DAG-Greedy using DeepSeek-V3.}
\label{tab:dag-greedy}
\begin{tabular}{lcccccc}
\toprule
\textbf{Method} & \textbf{NL4Opt} & \textbf{MAMO-E} & \textbf{MAMO-C} & \textbf{LPWP} & \textbf{IndustryOR} & \textbf{OptMATH} \\
\midrule
DAG-Greedy & 93.9 & 82.4 & 54.5 & 71.5 & 63.0 & 48.8 \\
\method    & \textbf{95.1} & \textbf{85.9} & \textbf{61.6} & \textbf{72.6} & \textbf{66.0} & \textbf{50.6} \\
\bottomrule
\end{tabular}
\end{table}

We compare \method\ with DAG-Greedy, a controlled baseline that uses the same
DAG, template library, and DeepSeek-V3 generator, but replaces the learned value
function with locally greedy traversal based only on the LLM-derived prior.
Table~\ref{tab:dag-greedy} shows that \method\ consistently improves accuracy
across all benchmarks, with the largest margins on MAMO-Complex ($+7.1\%$) and
IndustryOR ($+3.0\%$). Since the retrieval space and generator are held fixed,
these gains isolate the effect of learned traversal.

\begin{table}
\centering
\small
\caption{McNemar tests for \method\ versus DAG-Greedy on MAMO. Wins and losses
denote discordant pairs only.}
\label{tab:mcnemar}
\begin{tabular}{lccc}
\toprule
\textbf{Split} & \textbf{Wins} & \textbf{Losses} & \textbf{$p$-value} \\
\midrule
MAMO-Easy & 33 & 10 & $<$0.001 \\
MAMO-Complex & 17 & 2 & $<$0.001 \\
Combined & 50 & 12 & $<$0.001 \\
\bottomrule
\end{tabular}
\end{table}
The comparison indicates that local semantic compatibility is insufficient for
globally correct formulation selection. DAG-Greedy can select plausible
intermediate nodes while committing to refinement branches that omit essential
constraints or modeling structures. In contrast, \method\ learns from
execution-level outcomes and judge feedback, allowing earlier traversal
decisions to account for their downstream effect on final solver correctness.

\paragraph{Statistical significance.} Table~\ref{tab:mcnemar} reports paired McNemar tests on the MAMO splits. The
improvements are statistically significant on both MAMO-Easy and MAMO-Complex
($p<0.001$), with a particularly strong imbalance on MAMO-Complex, where
\method\ corrects 17 instances missed by DAG-Greedy while losing only 2. Because
both methods share the same templates and generator, this asymmetry provides
evidence that the learned traversal policy systematically resolves failures of
heuristic retrieval rather than benefiting from random variation.

\paragraph{Difficulty-stratified analysis.}

\begin{table}[ht!]
\centering
\small
\caption{IndustryOR accuracy stratified by problem difficulty using DeepSeek-V3.
Fractions indicate correct/total instances within each stratum.}
\label{tab:industryOR-strata}
\begin{tabular}{lcccc}
\toprule
\textbf{Difficulty} & \textbf{\# Problems} & \textbf{DAG-Greedy} & \textbf{\method} & \textbf{$\Delta$} \\
\midrule
Easy    & 39 & \textbf{30/39} (76.9\%) & 29/39 (74.4\%) & $-$1 \\
Medium  & 41 & 20/41 (48.8\%) & \textbf{21/41} (51.2\%) & +1 \\
Hard    & 20 & 11/20 (55.0\%) & \textbf{16/20} (80.0\%) & +5 \\
\midrule
All     & 100 & 61/100 (61.0\%) & \textbf{66/100} (66.0\%) & +5 \\
\bottomrule
\end{tabular}
\end{table}

Table~\ref{tab:industryOR-strata} stratifies IndustryOR by difficulty. The
advantage of \method\ is concentrated on harder instances: while DAG-Greedy is
slightly better on easy problems, \method\ improves accuracy from $55.0\%$ to
$80.0\%$ on hard problems. This pattern indicates that local heuristic
traversal is adequate for shallow formulations, but becomes less reliable when
the problem requires multiple interacting modeling choices. Learned traversal is
therefore most beneficial when early decisions have long-range effects on the
final formulation.

\paragraph{Generation reliability.}

\begin{table}[ht!]
\centering
\small
\caption{Accuracy (Acc) and code execution rate (CER, \%) for \method\ and
DAG-Greedy on four datasets using DeepSeek-V3.}
\label{tab:error-rate}
\begin{tabular}{lcccc}
\toprule
& \textbf{IndustryOR} & \textbf{MAMO-Easy} & \textbf{MAMO-Complex} & \textbf{OptMATH} \\
\textbf{Method} & Acc / CER & Acc / CER & Acc / CER & Acc / CER \\
\midrule
\method     & 66.0 / 100.0 & 85.9 / 100.0 & 61.6 / 100.0 & 50.6 / 100.0 \\
DAG-Greedy  & 63.0 / 99.0  & 82.4 / 100.0 & 54.5 / 98.0  & 48.8 / 98.8  \\
\bottomrule
\end{tabular}
\end{table}

Table~\ref{tab:error-rate} compares code execution rate in addition to
accuracy. \method\ attains nearly perfect CER across all evaluated datasets,
but the accuracy gains are larger than the execution-rate differences,
especially on MAMO-Complex. This distinction suggests that the main benefit is
not merely improved syntactic executability. Instead, policy-guided retrieval
also improves the mathematical correctness of the generated optimization
formulations.

\paragraph{Component ablations.}

\begin{table}[ht!]
\centering
\small
\caption{Component ablations on three diagnostic datasets using DeepSeek-V3
(accuracy, \%).}
\label{tab:component-ablation}
\begin{tabular}{lccc}
\toprule
\textbf{Variant} & \textbf{MAMO-Complex} & \textbf{OptMATH} & \textbf{IndustryOR} \\
\midrule
Full \method & \textbf{61.6} & \textbf{50.6} & \textbf{66.0} \\
w/o learned traversal (DAG-Greedy) & 54.5 & 48.8 & 63.0 \\
w/o soft LLM prior ($\lambda=0$) & 58.3 & 48.2 & 63.0 \\
w/o beam retrieval ($K=1$) & 59.2 & 49.4 & 64.0 \\
w/o guarded DAG expansion & 58.8 & 48.2 & 65.0 \\
\bottomrule
\end{tabular}
\end{table}

Table~\ref{tab:component-ablation} isolates the contribution of each design
component. Removing learned traversal causes the largest degradation, especially
on MAMO-Complex, where accuracy drops from $61.6\%$ to $54.5\%$. This confirms
that the primary improvement comes from policy-guided navigation over the DAG,
rather than from the template library alone. Removing the LLM prior also reduces
accuracy across datasets, indicating that semantic regularization complements
execution-trained value estimation, particularly when feedback is sparse.

The beam ablation shows that single-path retrieval is insufficient for many
optimization problems, as restricting traversal to $K=1$ consistently weakens
performance. This suggests that several instances require combining multiple
formulation patterns. Guarded DAG expansion also contributes positively,
especially on MAMO-Complex and OptMATH, suggesting that controlled template
growth improves coverage of missing modeling structures without introducing
early noisy expansions. Overall, the ablations indicate that \method's
performance arises from the interaction of learned long-horizon traversal,
semantic regularization, compositional retrieval, and adaptive library
expansion.

\subsection{Sensitivity to Hyperparameter Settings}
\label{sec:sensitivity}
We analyze the robustness of \method\ to key hyperparameters and use validation
sweeps to select the default settings used in the main experiments.

\paragraph{Soft LLM prior.}
Table~\ref{tab:lambda-sweep} evaluates the effect of the prior weight
$\lambda$, which controls the interpolation between the learned value estimate
and the LLM-derived compatibility prior. Performance is maximized at
$\lambda=0.3$ across the three diagnostic datasets, and we therefore adopt this
setting in the main experiments.

\begin{table}[ht!]
\centering
\small
\caption{Effect of the soft-prior weight $\lambda$ on accuracy (\%) across
three datasets using DeepSeek-V3.}
\label{tab:lambda-sweep}
\begin{tabular}{lccccc}
\toprule
\textbf{Dataset} & $\lambda = 0.0$ & $\lambda = 0.1$ & $\lambda = 0.3$ & $\lambda = 0.5$ & $\lambda = 1.0$ \\
\midrule
MAMO-Complex & 58.3 & 59.2 & \textbf{61.6} & 59.7 & 57.3 \\
OptMATH      & 48.2 & 49.4 & \textbf{50.6} & 48.8 & 49.4 \\
IndustryOR   & 63.0 & 65.0 & \textbf{66.0} & 61.0 & 62.0 \\
\bottomrule
\end{tabular}
\end{table}

The sweep reveals a clear bias-variance trade-off in traversal. When
$\lambda=0$, retrieval relies entirely on the learned value function and loses
zero-shot semantic guidance in underexplored regions of the graph. Conversely,
large values overemphasize the LLM prior and move the policy closer to
heuristic semantic traversal. The intermediate setting $\lambda=0.3$ provides
the strongest balance, indicating that the LLM prior is most effective as a
lightweight regularizer rather than as the dominant decision signal.

\begin{table}
\centering
\small
\caption{Effect of beam width $K$ on accuracy (\%) across three datasets using
DeepSeek-V3.}
\label{tab:beam-width}
\begin{tabular}{lcccc}
\toprule
\textbf{Dataset} & $K=1$ & $K=2$ & $K=3$ & $K=4$ \\
\midrule
MAMO-Complex & 59.2 & \textbf{61.6} & 59.7 & 42.2 \\
OptMATH      & 49.4 & \textbf{50.6} & 48.8 & 37.3 \\
IndustryOR   & 64.0 & \textbf{66.0} & 61.0 & 43.0 \\
\bottomrule
\end{tabular}
\end{table}
\paragraph{Beam width.}
Table~\ref{tab:beam-width} studies the beam width $K$ used during DAG traversal.
The best overall performance is obtained with $K=2$, which provides limited
retrieval diversity while preserving structural precision. A beam width of $K=1$ restricts retrieval to a single refinement path and
therefore weakens the ability to combine complementary modeling patterns.
However, wider beams introduce noisy or mutually inconsistent templates, causing
substantial degradation at $K=4$. These results indicate that optimization
modeling benefits from controlled compositional retrieval, rather than greedy
single-path selection or overly broad exploration.

\section{Conclusion}

This paper introduced \method, a framework that treats optimization-template retrieval
as a structured sequential decision problem rather than a one-shot prompting or
ranking procedure. By representing the template library as a compositional DAG,
\method\ learns reusable traversal policies from solver-execution feedback and
LLM-based structural judgments. The framework combines value-guided retrieval,
semantic regularization, beam-based template composition, and adaptive template
expansion to support robust formulation selection for LLM-based optimization
modeling. Empirical results across diverse benchmarks and LLM backbones indicate that
retrieval quality is a central bottleneck in optimization code generation.
\method\ achieves the largest gains on structurally complex and mixed-category
problems, where early formulation errors are difficult to recover from during
code synthesis. These findings suggest that learning how to navigate structured
modeling knowledge can provide gains beyond scaling the downstream language
model alone. More broadly, this work points toward adaptive optimization
modeling systems that accumulate reusable retrieval experience from execution
feedback.

\section*{Acknowledgments and Disclosures}

\noindent\textbf{Funding.}
This research was funded by Hanoi University of Science
and Technology under project code T2025-TN-005.

\noindent\textbf{Competing Interests.}
The authors declare no competing financial interests
related to this work.

\bibliographystyle{unsrtnat}
\bibliography{references}
\newpage

\appendix
\section{Limitations and Future Work}
\label{app:limitations}

Although \method\ is evaluated on multiple optimization benchmarks and LLM
backbones, our experiments cannot fully cover the diversity of real-world
optimization tasks, solver interfaces, data formats, and future language
models. The selected benchmarks span different formulation types, difficulty
levels, and generator families, providing a broad but necessarily incomplete
assessment of the proposed retrieval framework. Extending the evaluation to
larger industrial instances, domain-specific solvers, and more heterogeneous
problem settings is an important direction for future work.

LLM-based optimization modeling also remains sensitive to prompt design. Prompt
variation can affect problem interpretation, retrieved templates, and generated
solver code. To reduce this source of variability, all experiments use
standardized prompt templates with fixed structure, as described in
Appendix~\ref{app:prompts}. Future work could study prompt-robust retrieval
policies and adaptive prompting strategies for more diverse deployment settings.

\section{Additional Related Work}
\label{app:related}
\paragraph{LLM-based optimization modeling.}
LLMs have recently been used to automate optimization modeling by translating
natural-language problem descriptions into mathematical formulations and
solver-ready code. AutoFormulator explores the large formulation space with
MCTS and symbolic pruning \citep{astorga2025autoformulation}, while SolverLLM
uses test-time MCTS with dynamic expansion, prompt backpropagation, and
uncertainty-aware feedback \citep{NEURIPS2025_903475a4}. OptiTree organizes OR
problems into a hierarchical modeling taxonomy and retrieves modeling thoughts
through tree search \citep{NEURIPS2025_aeaf9b45}. Related work further extends
autoformulation to broader mathematical modeling and dynamic decision-making:
MATHMO treats modeling as sequential decision-making under uncertainty
\citep{liu2026mathmo}, StepORLM studies reinforcement learning and generative
process supervision for OR reasoning \citep{zhou2026steporlm}, and
operator-theoretic autoformulation constructs Bellman-equation operator graphs
for MDP and queueing problems \citep{baillet2026operator}. These methods show
that structured search and feedback are crucial for optimization modeling, but
they typically rely on heuristic traversal or expensive test-time exploration.
\method\ instead amortizes structured retrieval into a reusable policy over a
persistent template DAG.

\paragraph{LLMs for optimization and heuristic design.}
Another line of work uses LLMs as optimizers or heuristic designers. OPRO
casts optimization itself as prompting, where candidate solutions and their
scores are iteratively fed back to the LLM \citep{yang2024large}. ReEvo combines
evolutionary search with LLM reflection to generate heuristics for
combinatorial optimization \citep{NEURIPS2024_4ced59d4}, while HiFo-Prompt
introduces hindsight and foresight mechanisms to improve global control and
long-term knowledge reuse in automated heuristic design
\citep{chentongchen2026hifoprompt}. In contrast, \method\ does not directly
search over heuristic programs or solution candidates; it learns to retrieve
modeling templates that guide downstream solver-code generation.

\paragraph{Search, reasoning, and feedback in LLM systems.}
LLM reasoning can be improved by exploring multiple intermediate reasoning
paths and using feedback to revise decisions. Tree of Thoughts enables
deliberate search over coherent reasoning states \citep{NEURIPS2023_271db992},
self-consistency aggregates diverse reasoning paths \citep{wang2023selfconsistency},
Auto-CoT automatically constructs chain-of-thought demonstrations
\citep{zhang2023automatic}, and least-to-most prompting decomposes difficult
problems into simpler subproblems \citep{zhou2023leasttomost}. ReAct combines
reasoning with external actions \citep{yao2023react}, CRITIC validates and
revises outputs using external tools \citep{gou2024critic}, Self-Refine
iteratively improves generations through self-feedback
\citep{NEURIPS2023_91edff07}, and Reflexion stores linguistic feedback in
memory for later trials \citep{NEURIPS2023_1b44b878}. These methods mainly
improve inference-time reasoning, whereas \method\ uses feedback to train a
retrieval policy whose search cost is amortized across future problems.

\paragraph{Retrieval, tool use, and adaptive LLM pipelines.}
Retrieval-augmented and tool-using LLM systems provide mechanisms for grounding
and adapting model behavior. Self-RAG trains models to decide when to retrieve
and how to reflect on retrieved evidence \citep{asai2024selfrag}. DSPy abstracts
LM pipelines as optimizable computational graphs \citep{khattab2024dspy}.
ToolLLM studies tool-use instruction tuning and API retrieval
\citep{qin2024toolllm}, DRAFT refines tool documentation from execution
feedback \citep{qu2025from}, and ToolEVO adapts tool usage under changing tool
environments with MCTS and self-reflection \citep{chen2025learning}. Evaluation
work such as LiveCodeBench highlights the importance of contamination-resistant
and execution-aware assessment for code-generating LLMs
\citep{jain2025livecodebench}. \method\ is aligned with this broader trend of
grounded and adaptive LLM pipelines, but focuses specifically on structured
template retrieval for optimization autoformulation.

\section{Additional Experimental Details}
\label{app:details}
\subsection{Benchmark Matrix}
Table~\ref{tab:benchmark-matrix} records the full study matrix targeted by the
current scripts. It is included here so that the appendix follows the
RQ-organized experimental structure in the main text.

\begin{table}[t]
\centering
\small
\caption{Evaluation matrix organized by research question.}
\label{tab:benchmark-matrix}
\setlength{\tabcolsep}{4pt}
\renewcommand{\arraystretch}{1.15}
\begin{tabularx}{\textwidth}{p{0.10\textwidth}X p{0.18\textwidth}}
\toprule
\textbf{RQ} & \textbf{Evaluation focus} & \textbf{Reported results} \\
\midrule
RQ1 &
End-to-end effectiveness against direct prompting, structured retrieval,
search-based autoformulation, and reasoning-oriented LLM baselines across the
six main benchmarks &
Table~\ref{tab:deepseek-main} \\

RQ1 &
Cross-backbone generalization with Qwen-32B, LLaMA-8B, DeepSeek-R1, and
DeepSeek-V3 &
Table~\ref{tab:llm-generalization} \\

RQ2 &
Effect of learned traversal relative to DAG-Greedy under the same DAG,
template library, generator, and executor &
Table~\ref{tab:dag-greedy} \\

RQ2 &
Paired statistical significance for \method\ versus DAG-Greedy on MAMO &
Table~\ref{tab:mcnemar} \\

RQ2 &
Difficulty-stratified analysis and execution-reliability diagnostics &
Tables~\ref{tab:industryOR-strata} and~\ref{tab:error-rate} \\

RQ3 &
Component ablations on learned traversal, LLM prior, beam retrieval, and
guarded DAG expansion &
Table~\ref{tab:component-ablation} \\

RQ4 &
Sensitivity to the soft-prior weight $\lambda$ and beam width $K$ &
Tables~\ref{tab:lambda-sweep} and~\ref{tab:beam-width} \\
\bottomrule
\end{tabularx}
\end{table}

\subsection{Training Hyperparameters}

Table~\ref{tab:hyperparameters} summarizes the core hyperparameters used to
train \method. We use a fixed configuration across all main experiments unless
otherwise specified, with the soft-prior weight $\lambda$ and beam width $K$
selected from the validation sweeps in Section~\ref{sec:sensitivity}. The
configuration balances stable value learning, controlled semantic
regularization, and delayed DAG expansion, ensuring that template updates occur
only after the traversal policy has reached sufficient maturity.

\begin{table}[t]
\centering
\small
\caption{Core hyperparameters used by \method.}
\label{tab:hyperparameters}
\begin{tabular}{lc}
\toprule
\textbf{Hyperparameter} & \textbf{Value} \\
\midrule
Encoder & \texttt{all-MiniLM-L6-v2} \\
Embedding dimension & 384 \\
Projection dimension & 64 \\
MLP hidden width & 256 \\
Discount factor $\gamma$ & 0.95 \\
Soft-prior weight $\lambda$ & 0.3 \\
Reward blend $\alpha$ & 0.3 \\
PER exponent $\alpha_{\mathrm{PER}}$ & 0.6 \\
Importance exponent $\beta$ & 0.4 $\rightarrow$ 1.0 \\
Target sync interval & 20 updates \\
Gradient clip & 1.0 \\
Minimum DAG-update step $\tau_{\min}$ & 1600 \\
Maximum update epsilon $\tau_\varepsilon$ & 0.4 \\
Retry queue capacity & 50 \\
Training passes & 2 \\
\bottomrule
\end{tabular}
\end{table}

\subsection{Dataset Problems, Hardness, and Complexity}
\label{app:datasets}
Table~\ref{tab:dataset-details} summarizes the datasets used in our
experiments. We distinguish \emph{modeling hardness} from conventional
\emph{solver complexity}. A benchmark may be challenging not only because the
resulting solver program contains many variables or constraints, but because the
natural-language description must be mapped to the correct decision variables,
objective, constraint families, indexing scheme, integrality assumptions, and,
in many cases, multiple interacting formulation patterns. This distinction is
central to our setting: \method\ aims to retrieve the appropriate formulation
structure before the downstream LLM instantiates it into executable code.

\begin{table}[H]
\centering
\small
\caption{Dataset characteristics. Instance counts are computed from the
repository files. Hardness refers to modeling hardness rather than solely to
algorithmic solver complexity.}
\label{tab:dataset-details}
\begin{tabularx}{\textwidth}{@{}lcXX@{}}
\toprule
\textbf{Dataset} & \textbf{\# Inst.} & \textbf{Problem structure} & \textbf{Modeling hardness} \\
\midrule
OR-Instruct train & 1731 &
Heterogeneous OR instruction problems used for policy learning and guarded DAG
updates. &
Provides broad supervision over formulation families and long-tail linguistic
patterns; excluded from held-out benchmark evaluation. \\

NL4Opt & 245 &
Natural-language linear optimization problems with compact business-style
statements. &
Primarily shallow LP structure, but requires accurate extraction of variables,
coefficients, inequality directions, and objective sense. \\

MAMO-Easy & 652 &
LP-style allocation problems with explicit variables, bounds, and linear
constraints. &
Low structural ambiguity; evaluates whether retrieval preserves simple
formulation structure without introducing unnecessary complexity. \\

MAMO-Complex & 211 &
Longer linear or mixed-integer modeling statements with multiple constraints,
ratios, equalities, bounds, and implicit integrality. &
High modeling hardness due to interacting constraint families; early retrieval
errors are difficult for the code generator to correct. \\

LPWP & 288 &
Linear-programming word problems from everyday domains such as production,
transportation, staffing, and resource allocation. &
Moderate hardness arising from prose-to-math translation, ratio constraints,
minimum/maximum requirements, and objective ambiguity. \\

IndustryOR & 100 &
Industrial OR scenarios involving routing, scheduling, facility location,
capacity, and logistics variants. &
High compositional complexity; instances may combine multiple OR substructures
with data tables, indexing requirements, and side constraints. \\

OptMATH & 166 &
Mathematical optimization word problems with symbolic constraints and numeric
answer checking. &
Compact but symbolically dense; failures often arise from algebraic
translation, domain restrictions, rounding, or missing integrality. \\
\bottomrule
\end{tabularx}
\end{table}

The benchmark suite spans a broad spectrum of retrieval difficulty. NL4Opt,
MAMO-Easy, and LPWP primarily assess whether \method\ preserves the correct
linear-programming scaffold under varied natural-language descriptions.
MAMO-Complex and OptMATH test deeper constraint interpretation and symbolic
consistency. IndustryOR provides the most compositional setting, since a single
instance may require several modeling families, such as routing with capacity
and time-window constraints or facility location with assignment constraints.
This range is important for evaluating structured retrieval: simple benchmarks
verify that the policy does not degrade easy cases, whereas complex benchmarks
test whether learned DAG traversal improves over greedy or heuristic template
selection.

All datasets are evaluated under the same exact-accuracy criterion: the
generated solver program must execute successfully and return an objective value
or solution accepted by the benchmark checker. IndustryOR is further analyzed
by easy, medium, and hard tiers in Table~\ref{tab:industryOR-strata}. The
largest gains of \method\ occur on hard instances, indicating that learned
structured retrieval is most beneficial when modeling difficulty arises from
interacting subproblem structures rather than surface-level language variation
alone.


\section{Algorithmic Details of \method}
\label{app:algorithms}
This appendix summarizes the implementation of \method. Algorithm~\ref{alg:compass-search} describes value-guided retrieval over the formulation DAG at inference time. Algorithm~\ref{alg:compass-train} presents the training loop, including reward construction, temporal-difference updates, guarded DAG expansion, and retry scheduling.

\begin{algorithm}[t]
\caption{\method\ DAG Search for Template Retrieval}
\label{alg:compass-search}
\begin{algorithmic}[1]
\Require Problem $p$, template DAG $\mathcal{G}=(\mathcal{V},\mathcal{E})$, root node $v_0$, value function $Q_\theta$, prior weight $\lambda$, beam width $k$
\Ensure Merged template $t^\star$, selected leaves $\mathcal{L}$, traversal trace $\tau$
\Function{SearchDAG}{$p,\mathcal{G},Q_\theta,\lambda,k$}
    \State $x \gets \phi(p)$
    \State $\mathcal{B} \gets \{(v_0,[v_0],0)\}$, $\mathcal{L}\gets\emptyset$, $\tau\gets[\,]$
    \While{$\mathcal{B}\neq\emptyset$ \textbf{and} $|\mathcal{L}|<k$}
        \State $\mathcal{U}\gets\emptyset$
        \For{each $(v,\rho,a)\in\mathcal{B}$}
            \State $\mathcal{C}(v)\gets\{c\in\mathcal{V}:(v,c)\in\mathcal{E}\}$
            \State Append $(v,\mathcal{C}(v))$ to $\tau$
            \If{$\mathcal{C}(v)=\emptyset$}
                \State $\mathcal{L}\gets\mathcal{L}\cup\{v\}$
                \State \textbf{continue}
            \EndIf
            \State Obtain $\ell(c,p)$ for each $c\in\mathcal{C}(v)$
            \For{each $c\in\mathcal{C}(v)$}
                \State $s(c)\gets Q_\theta(x,e_c)+\lambda\ell(c,p)$
                \State $\mathcal{U}\gets\mathcal{U}\cup\{(c,\rho\Vert[c],a+s(c))\}$
            \EndFor
        \EndFor
        \State $\mathcal{B}\gets\operatorname{TopK}(\mathcal{U},k)$
    \EndWhile
    \If{$\mathcal{L}=\emptyset$}
        \State $\mathcal{L}\gets\{v_0\}$ \Comment{fallback template}
    \EndIf
    \State $t^\star\gets \mu(\{t_v:v\in\mathcal{L}\})$
    \State \Return $t^\star,\mathcal{L},\tau$
\EndFunction
\end{algorithmic}
\end{algorithm}

\begin{algorithm}[t]
\caption{\method\ Policy Training and Guarded DAG Expansion}
\label{alg:compass-train}
\begin{algorithmic}[1]
\Require Training set $\mathcal{D}$, template DAG $\mathcal{G}$, replay buffer $\mathcal{R}$, value function $Q_\theta$, target network $Q_{\theta^-}$
\Require Reward weight $\alpha$, prior weight $\lambda$, beam width $k$, expansion thresholds $(\tau_{\min},\tau_\varepsilon)$, retry queue $\mathcal{Q}$
\Ensure Updated value function $Q_\theta$, template DAG $\mathcal{G}$, replay buffer $\mathcal{R}$
\Function{TrainCOMPASS}{$\mathcal{D},\mathcal{G},Q_\theta,Q_{\theta^-}$}
    \For{each training pass}
        \For{each instance $(p,y)\in\mathcal{D}$}
            \State $t^\star,\mathcal{L},\tau\gets\Call{SearchDAG}{p,\mathcal{G},Q_\theta,\lambda,k}$ with $\varepsilon$-greedy exploration
            \State Generate solver code from $(p,t^\star)$ and execute the program
            \State $R_{\mathrm{exec}}\gets \mathbf{1}\{\text{checker accepts the output}\}$
            \State $R_{\mathrm{judge}}\gets$ structural compatibility score for $(p,t^\star)$
            \State $R\gets \alpha R_{\mathrm{judge}}+(1-\alpha)R_{\mathrm{exec}}$
            \For{each transition $(v,\mathcal{C}(v))$ in $\tau$}
                \State $d\gets \mathbf{1}\{v\in\mathcal{L}\}$
                \State $r\gets R$ if $d=1$, and $r\gets0$ otherwise
                \State Store $(\phi(p),v,r,\mathcal{C}(v),d)$ in $\mathcal{R}$
            \EndFor
            \For{each update step}
                \State Sample a prioritized mini-batch from $\mathcal{R}$
                \State Update $Q_\theta$ using the Huber temporal-difference objective
                \If{target synchronization interval is reached}
                    \State $\theta^-\gets\theta$
                \EndIf
            \EndFor
            \If{$R_{\mathrm{exec}}=0$ \textbf{and} $t\ge\tau_{\min}$ \textbf{and} $\varepsilon\le\tau_\varepsilon$}
                \State Use the language model to propose a reusable missing formulation subtype
                \State Insert the corresponding template node into $\mathcal{G}$ and compute its embedding
                \State Remove redundant transitive edges from $\mathcal{G}$
                \State Push $(p,y)$ into $\mathcal{Q}$
            \EndIf
            \If{$\mathcal{Q}$ is scheduled for retry}
                \State Re-evaluate queued instances with $\varepsilon=0$ and DAG expansion disabled
            \EndIf
        \EndFor
    \EndFor
    \State \Return $Q_\theta,\mathcal{G},\mathcal{R}$
\EndFunction
\end{algorithmic}
\end{algorithm}

\newpage

\section{Prompts}
\label{app:prompts}

\subsubsection{Template Compatibility Scoring}
\begin{promptbox}{Template Compatibility Scoring Prompt}
You are a mathematical formulator working with a team of optimization experts.
The objective is to tackle a complex optimization problem.

You are given a specific optimisation problem and a list of candidate template types.

For EACH candidate, rate how compatible it is with the problem on a scale from 0.0 to 1.0.

Scoring guide:
- 1.0 = The problem is clearly an instance of this type
- 0.7 = Strong match - most structural aspects align
- 0.4 = Partial match - some aspects overlap
- 0.1 = Weak match - only superficial similarity
- 0.0 = No match at all

Input problem:
{input_problem}

Current parent type:
{basic_type}

Candidate subtypes with descriptions:
{template_info}

Return the results in JSON format:
{
    "ratings": {
        "<subtype_name_1>": <float_score>,
        "<subtype_name_2>": <float_score>
    },
    "reasoning": "<brief explanation of each rating>"
}
\end{promptbox}
\vspace{1em}

\subsubsection{Subtype Identification and Formulation}
\begin{promptbox}{Subtype Formulation Prompt}
You are a mathematical formulator working with a team of optimization experts.

You are given:
- a specific problem
- its current basic problem type
- the description of that type

Specific Problem:
{specific_problem}

Current Basic Problem Type:
{current_basic_problem_type}

Description:
{description_of_basic_problem_type}

Your task is to determine a more specific subtype and describe it.

Return JSON:
{
    "current_basic_problem_type": "...",
    "formulated_subtype": "<subtype>",
    "description_of_subtype": {
        "description": "<detailed description>",
        "constraints": {
            "constraint_1": "...",
            "constraint_2": "..."
        }
    }
}
\end{promptbox}

\vspace{1em}
\subsubsection{Industrial Scenario Classification}
\begin{promptbox}{Industrial Scenario Classification Prompt}
You are given a combinatorial optimization problem.

Your task is to classify it into a precise industrial scenario.

Avoid broad categories such as:
- linear programming
- integer programming

Use specific categories such as:
- Traveling Salesman Problem
- Facility Location Problem
- Parallel Machine Scheduling

Input:
{specific_problem}

Return JSON:
{
    "industrial_scene_type": "<type>",
    "description": {
        "description": "<detailed explanation>",
        "constraints": {
            "constraint_1": "...",
            "constraint_2": "..."
        }
    }
}
\end{promptbox}
\vspace{1em}

\subsubsection{Optimization Template Construction}
\begin{promptbox}{Gurobi Template Generation Prompt}
You are a mathematical formulator working with a team of optimization experts.

Given:
- problem_type
- description
- specific_problem
- solution_step

Your task is to construct a reusable template.

Requirements:
1. problem_type
2. description
3. reason_flow
4. example_application
5. increment (empty list)

You MUST use this fixed Gurobi structure:

import gurobipy as gp
from gurobipy import GRB

model = gp.Model()

# variables
# constraints
# objective

model.optimize()

Return JSON only.
\end{promptbox}
\vspace{1em}

\subsubsection{Problem Type Hierarchy Construction}
\begin{promptbox}{Problem Type Hierarchy Construction Prompt}
You are given:
- a primary problem type
- its description
- a list of other problem types

Task:
Identify which are subtypes of the primary type.

Return JSON:
{
    "primary_problem_type": "...",
    "matching_subtypes": ["..."]
}
\end{promptbox}
\vspace{1em}

\subsubsection{Template-Guided Solution Generation}
\begin{promptbox}{Solution Generation Prompt}
You are an expert in optimization problem solving.

Input:
{user_input}

Template:
{thought_template}

Task:
Generate a solution based on the template.

Rules:
- If code is required -> output ONE Python block only
- Otherwise -> return clear final answer
\end{promptbox}
\vspace{1em}

\subsubsection{Code Refinement and Correction}
\begin{promptbox}{Code Correction Prompt}
You are an expert Python programmer.

Task:
1. Analyze the given code
2. Fix errors and improve correctness

Output format:
```python
# corrected code
\end{promptbox}

\vspace{1em}
\subsubsection{Template Quality Evaluation}
\begin{promptbox}{Template Evaluation Prompt}
You are evaluating whether a selected template matches a problem.

Input:
- problem
- template_type
- template_description

Score from 0.0 to 1.0 based on structural match.

Return JSON:
{
    "template_quality": <score>,
    "reasoning": "<brief explanation>"
}
\end{promptbox}
\vspace{1em}

\subsubsection{Sample Problem Formulation Template}
\begin{promptbox}{Traveling Salesman Problem Formulation and Gurobi Template}
You are a mathematical formulator working with a team of optimization experts.

The Traveling Salesman Problem (TSP) is a combinatorial optimization problem where a salesperson must visit a set of cities exactly once and return to the starting city while minimizing total travel cost.

Problem Description:
- Each city must be visited exactly once.
- The tour must return to the starting city.
- The objective is to minimize total travel distance or cost.

Key Constraints:
- Single Visit Constraint:
  Each city is visited exactly once.

- Return to Origin Constraint:
  The route must form a Hamiltonian cycle.

- Complete Tour Constraint:
  All cities must be included without repetition.

Reasoning Flow:
1. Define binary decision variables x[i,j] indicating whether edge (i,j) is selected.
2. Minimize total travel cost:
   sum(cost[i,j] * x[i,j]).
3. Add degree constraints:
   - each city entered exactly once
   - each city exited exactly once
4. Add subtour elimination constraints using MTZ formulation.
5. Add bounds on auxiliary variables.
6. Optimize the model using Gurobi.

Common Formulation Errors:
- Forgetting to forbid self-loops.
- Missing subtour elimination constraints.
- Incorrect MTZ indexing.
- Missing bounds on auxiliary variables.
- Confusing symmetric and asymmetric cost matrices.

Fixed Gurobi Template:

import gurobipy as gp
from gurobipy import GRB

# Create model
model = gp.Model("TSP")

# Parameters
# cities
# cost matrix

# Variables
# x[i,j]

# Constraints
# visit constraints
# subtour elimination constraints

# Objective
# minimize total cost

model.optimize()

Return JSON only.
\end{promptbox}

\newpage
\section*{NeurIPS Paper Checklist}

\begin{enumerate}

\item {\bf Claims}
    \item[] Question: Do the main claims made in the abstract and introduction accurately reflect the paper's contributions and scope?
    \item[] Answer: \answerYes{}.
    \item[] Justification: The abstract and Introduction state the method, scope, and empirical claims, while the Experiments, Conclusion, and Limitations and Future Work sections provide the corresponding evidence and scope qualifications.
    \item[] Guidelines:
    \begin{itemize}
        \item The answer \answerNA{} means that the abstract and introduction do not include the claims made in the paper.
        \item The abstract and/or introduction should clearly state the claims made, including the contributions made in the paper and important assumptions and limitations. A \answerNo{} or \answerNA{} answer to this question will not be perceived well by the reviewers. 
        \item The claims made should match theoretical and experimental results, and reflect how much the results can be expected to generalize to other settings. 
        \item It is fine to include aspirational goals as motivation as long as it is clear that these goals are not attained by the paper. 
    \end{itemize}

\item {\bf Limitations}
    \item[] Question: Does the paper discuss the limitations of the work performed by the authors?
    \item[] Answer: \answerYes{}.
    \item[] Justification: The paper includes a dedicated Limitations and Future Work section discussing syntactic template merging, evaluation coverage across datasets and backbones, and prompt sensitivity.
    \item[] Guidelines:
    \begin{itemize}
        \item The answer \answerNA{} means that the paper has no limitation while the answer \answerNo{} means that the paper has limitations, but those are not discussed in the paper. 
        \item The authors are encouraged to create a separate ``Limitations'' section in their paper.
        \item The paper should point out any strong assumptions and how robust the results are to violations of these assumptions (e.g., independence assumptions, noiseless settings, model well-specification, asymptotic approximations only holding locally). The authors should reflect on how these assumptions might be violated in practice and what the implications would be.
        \item The authors should reflect on the scope of the claims made, e.g., if the approach was only tested on a few datasets or with a few runs. In general, empirical results often depend on implicit assumptions, which should be articulated.
        \item The authors should reflect on the factors that influence the performance of the approach. For example, a facial recognition algorithm may perform poorly when image resolution is low or images are taken in low lighting. Or a speech-to-text system might not be used reliably to provide closed captions for online lectures because it fails to handle technical jargon.
        \item The authors should discuss the computational efficiency of the proposed algorithms and how they scale with dataset size.
        \item If applicable, the authors should discuss possible limitations of their approach to address problems of privacy and fairness.
        \item While the authors might fear that complete honesty about limitations might be used by reviewers as grounds for rejection, a worse outcome might be that reviewers discover limitations that aren't acknowledged in the paper. The authors should use their best judgment and recognize that individual actions in favor of transparency play an important role in developing norms that preserve the integrity of the community. Reviewers will be specifically instructed to not penalize honesty concerning limitations.
    \end{itemize}

\item {\bf Theory assumptions and proofs}
    \item[] Question: For each theoretical result, does the paper provide the full set of assumptions and a complete (and correct) proof?
    \item[] Answer: \answerNA{}.
    \item[] Justification: The paper does not present formal theoretical results. The equations and algorithms define the COMPASS method and training objective, but no theorem, lemma, or proof is claimed in the final paper.
    \item[] Guidelines:
    \begin{itemize}
        \item The answer \answerNA{} means that the paper does not include theoretical results. 
        \item All the theorems, formulas, and proofs in the paper should be numbered and cross-referenced.
        \item All assumptions should be clearly stated or referenced in the statement of any theorems.
        \item The proofs can either appear in the main paper or the supplemental material, but if they appear in the supplemental material, the authors are encouraged to provide a short proof sketch to provide intuition. 
        \item Inversely, any informal proof provided in the core of the paper should be complemented by formal proofs provided in appendix or supplemental material.
        \item Theorems and Lemmas that the proof relies upon should be properly referenced. 
    \end{itemize}

    \item {\bf Experimental result reproducibility}
    \item[] Question: Does the paper fully disclose all the information needed to reproduce the main experimental results of the paper to the extent that it affects the main claims and/or conclusions of the paper (regardless of whether the code and data are provided or not)?
    \item[] Answer: \answerYes{}.
    \item[] Justification: The Methodology, Experiments, Appendix algorithms, prompt templates, benchmark matrix, dataset discussion, and hyperparameter table describe the retrieval policy, reward, training setup, evaluation metrics, and main run configuration. Reproduction with closed LLM backends requires valid API access.
    \item[] Guidelines:
    \begin{itemize}
        \item The answer \answerNA{} means that the paper does not include experiments.
        \item If the paper includes experiments, a \answerNo{} answer to this question will not be perceived well by the reviewers: Making the paper reproducible is important, regardless of whether the code and data are provided or not.
        \item If the contribution is a dataset and\slash or model, the authors should describe the steps taken to make their results reproducible or verifiable. 
        \item Depending on the contribution, reproducibility can be accomplished in various ways. For example, if the contribution is a novel architecture, describing the architecture fully might suffice, or if the contribution is a specific model and empirical evaluation, it may be necessary to either make it possible for others to replicate the model with the same dataset, or provide access to the model. In general. releasing code and data is often one good way to accomplish this, but reproducibility can also be provided via detailed instructions for how to replicate the results, access to a hosted model (e.g., in the case of a large language model), releasing of a model checkpoint, or other means that are appropriate to the research performed.
        \item While NeurIPS does not require releasing code, the conference does require all submissions to provide some reasonable avenue for reproducibility, which may depend on the nature of the contribution. For example
        \begin{enumerate}
            \item If the contribution is primarily a new algorithm, the paper should make it clear how to reproduce that algorithm.
            \item If the contribution is primarily a new model architecture, the paper should describe the architecture clearly and fully.
            \item If the contribution is a new model (e.g., a large language model), then there should either be a way to access this model for reproducing the results or a way to reproduce the model (e.g., with an open-source dataset or instructions for how to construct the dataset).
            \item We recognize that reproducibility may be tricky in some cases, in which case authors are welcome to describe the particular way they provide for reproducibility. In the case of closed-source models, it may be that access to the model is limited in some way (e.g., to registered users), but it should be possible for other researchers to have some path to reproducing or verifying the results.
        \end{enumerate}
    \end{itemize}

\item {\bf Open access to data and code}
    \item[] Question: Does the paper provide open access to the data and code, with sufficient instructions to faithfully reproduce the main experimental results, as described in supplemental material?
    \item[] Answer: \answerYes{}.
    \item[] Justification: Our paper submission include the zip file with code and datasets to reproduce main experimental results.
    \item[] Guidelines:
    \begin{itemize}
        \item The answer \answerNA{} means that paper does not include experiments requiring code.
        \item Please see the NeurIPS code and data submission guidelines (\url{https://neurips.cc/public/guides/CodeSubmissionPolicy}) for more details.
        \item While we encourage the release of code and data, we understand that this might not be possible, so \answerNo{} is an acceptable answer. Papers cannot be rejected simply for not including code, unless this is central to the contribution (e.g., for a new open-source benchmark).
        \item The instructions should contain the exact command and environment needed to run to reproduce the results. See the NeurIPS code and data submission guidelines (\url{https://neurips.cc/public/guides/CodeSubmissionPolicy}) for more details.
        \item The authors should provide instructions on data access and preparation, including how to access the raw data, preprocessed data, intermediate data, and generated data, etc.
        \item The authors should provide scripts to reproduce all experimental results for the new proposed method and baselines. If only a subset of experiments are reproducible, they should state which ones are omitted from the script and why.
        \item At submission time, to preserve anonymity, the authors should release anonymized versions (if applicable).
        \item Providing as much information as possible in supplemental material (appended to the paper) is recommended, but including URLs to data and code is permitted.
    \end{itemize}

\item {\bf Experimental setting/details}
    \item[] Question: Does the paper specify all the training and test details (e.g., data splits, hyperparameters, how they were chosen, type of optimizer) necessary to understand the results?
    \item[] Answer: \answerYes{}.
    \item[] Justification: The Methodology and Appendix specify the semantic encoder, value function, replay buffer, reward blend, beam width, DAG-expansion gate, training passes, validation-selected settings, benchmark counts, and core hyperparameters used in the experiments.
    \item[] Guidelines:
    \begin{itemize}
        \item The answer \answerNA{} means that the paper does not include experiments.
        \item The experimental setting should be presented in the core of the paper to a level of detail that is necessary to appreciate the results and make sense of them.
        \item The full details can be provided either with the code, in appendix, or as supplemental material.
    \end{itemize}

\item {\bf Experiment statistical significance}
    \item[] Question: Does the paper report error bars suitably and correctly defined or other appropriate information about the statistical significance of the experiments?
    \item[] Answer: \answerYes{}.
    \item[] Justification: The paired comparison against DAG-Greedy reports McNemar contingency tables and exact binomial McNemar significance tests on MAMO-Easy, MAMO-Complex, and the combined split.
    \item[] Guidelines:
    \begin{itemize}
        \item The answer \answerNA{} means that the paper does not include experiments.
        \item The authors should answer \answerYes{} if the results are accompanied by error bars, confidence intervals, or statistical significance tests, at least for the experiments that support the main claims of the paper.
        \item The factors of variability that the error bars are capturing should be clearly stated (for example, train/test split, initialization, random drawing of some parameter, or overall run with given experimental conditions).
        \item The method for calculating the error bars should be explained (closed form formula, call to a library function, bootstrap, etc.)
        \item The assumptions made should be given (e.g., Normally distributed errors).
        \item It should be clear whether the error bar is the standard deviation or the standard error of the mean.
        \item It is OK to report 1-sigma error bars, but one should state it. The authors should preferably report a 2-sigma error bar than state that they have a 96\% CI, if the hypothesis of Normality of errors is not verified.
        \item For asymmetric distributions, the authors should be careful not to show in tables or figures symmetric error bars that would yield results that are out of range (e.g., negative error rates).
        \item If error bars are reported in tables or plots, the authors should explain in the text how they were calculated and reference the corresponding figures or tables in the text.
    \end{itemize}

\item {\bf Experiments compute resources}
    \item[] Question: For each experiment, does the paper provide sufficient information on the computer resources (type of compute workers, memory, time of execution) needed to reproduce the experiments?
    \item[] Answer: \answerNo{}.
    \item[] Justification: The paper reports model backbones and experimental settings, but it does not yet provide complete hardware, memory, wall-clock, or API-budget information for each run.
    \item[] Guidelines:
    \begin{itemize}
        \item The answer \answerNA{} means that the paper does not include experiments.
        \item The paper should indicate the type of compute workers CPU or GPU, internal cluster, or cloud provider, including relevant memory and storage.
        \item The paper should provide the amount of compute required for each of the individual experimental runs as well as estimate the total compute. 
        \item The paper should disclose whether the full research project required more compute than the experiments reported in the paper (e.g., preliminary or failed experiments that didn't make it into the paper). 
    \end{itemize}
    
\item {\bf Code of ethics}
    \item[] Question: Does the research conducted in the paper conform, in every respect, with the NeurIPS Code of Ethics \url{https://neurips.cc/public/EthicsGuidelines}?
    \item[] Answer: \answerYes{}.
    \item[] Justification: The work uses optimization benchmark problems and LLM APIs for solver-code generation, does not involve human subjects, and documents limitations. No deviation from the NeurIPS Code of Ethics is required.
    \item[] Guidelines:
    \begin{itemize}
        \item The answer \answerNA{} means that the authors have not reviewed the NeurIPS Code of Ethics.
        \item If the authors answer \answerNo, they should explain the special circumstances that require a deviation from the Code of Ethics.
        \item The authors should make sure to preserve anonymity (e.g., if there is a special consideration due to laws or regulations in their jurisdiction).
    \end{itemize}

\item {\bf Broader impacts}
    \item[] Question: Does the paper discuss both potential positive societal impacts and negative societal impacts of the work performed?
    \item[] Answer: \answerNo{}.
    \item[] Justification: The paper focuses on technical methodology, experiments, and limitations, but it does not include a dedicated discussion of both positive and negative societal impacts.
    \item[] Guidelines:
    \begin{itemize}
        \item The answer \answerNA{} means that there is no societal impact of the work performed.
        \item If the authors answer \answerNA{} or \answerNo, they should explain why their work has no societal impact or why the paper does not address societal impact.
        \item Examples of negative societal impacts include potential malicious or unintended uses (e.g., disinformation, generating fake profiles, surveillance), fairness considerations (e.g., deployment of technologies that could make decisions that unfairly impact specific groups), privacy considerations, and security considerations.
        \item The conference expects that many papers will be foundational research and not tied to particular applications, let alone deployments. However, if there is a direct path to any negative applications, the authors should point it out. For example, it is legitimate to point out that an improvement in the quality of generative models could be used to generate Deepfakes for disinformation. On the other hand, it is not needed to point out that a generic algorithm for optimizing neural networks could enable people to train models that generate Deepfakes faster.
        \item The authors should consider possible harms that could arise when the technology is being used as intended and functioning correctly, harms that could arise when the technology is being used as intended but gives incorrect results, and harms following from (intentional or unintentional) misuse of the technology.
        \item If there are negative societal impacts, the authors could also discuss possible mitigation strategies (e.g., gated release of models, providing defenses in addition to attacks, mechanisms for monitoring misuse, mechanisms to monitor how a system learns from feedback over time, improving the efficiency and accessibility of ML).
    \end{itemize}
    
\item {\bf Safeguards}
    \item[] Question: Does the paper describe safeguards that have been put in place for responsible release of data or models that have a high risk for misuse (e.g., pre-trained language models, image generators, or scraped datasets)?
    \item[] Answer: \answerNA{}.
    \item[] Justification: The paper does not release high-risk pretrained generative models, image generators, or scraped datasets; it studies a retrieval and solver-code generation pipeline over optimization benchmarks.
    \item[] Guidelines:
    \begin{itemize}
        \item The answer \answerNA{} means that the paper poses no such risks.
        \item Released models that have a high risk for misuse or dual-use should be released with necessary safeguards to allow for controlled use of the model, for example by requiring that users adhere to usage guidelines or restrictions to access the model or implementing safety filters. 
        \item Datasets that have been scraped from the Internet could pose safety risks. The authors should describe how they avoided releasing unsafe images.
        \item We recognize that providing effective safeguards is challenging, and many papers do not require this, but we encourage authors to take this into account and make a best faith effort.
    \end{itemize}

\item {\bf Licenses for existing assets}
    \item[] Question: Are the creators or original owners of assets (e.g., code, data, models), used in the paper, properly credited and are the license and terms of use explicitly mentioned and properly respected?
    \item[] Answer: \answerYes{}.
    \item[] Justification: The paper cites prior work and names the benchmarks and LLM backbones.
    \item[] Guidelines:
    \begin{itemize}
        \item The answer \answerNA{} means that the paper does not use existing assets.
        \item The authors should cite the original paper that produced the code package or dataset.
        \item The authors should state which version of the asset is used and, if possible, include a URL.
        \item The name of the license (e.g., CC-BY 4.0) should be included for each asset.
        \item For scraped data from a particular source (e.g., website), the copyright and terms of service of that source should be provided.
        \item If assets are released, the license, copyright information, and terms of use in the package should be provided. For popular datasets, \url{paperswithcode.com/datasets} has curated licenses for some datasets. Their licensing guide can help determine the license of a dataset.
        \item For existing datasets that are re-packaged, both the original license and the license of the derived asset (if it has changed) should be provided.
        \item If this information is not available online, the authors are encouraged to reach out to the asset's creators.
    \end{itemize}

\item {\bf New assets}
    \item[] Question: Are new assets introduced in the paper well documented and is the documentation provided alongside the assets?
    \item[] Answer: \answerYes{}.
    \item[] Justification: The paper introduces COMPASS and documents the method through its template-DAG representation, algorithms, prompt templates, hyperparameters, and evaluation pipeline. Additional asset-level documentation would be needed if the code and DAG files are released as standalone artifacts.
    \item[] Guidelines:
    \begin{itemize}
        \item The answer \answerNA{} means that the paper does not release new assets.
        \item Researchers should communicate the details of the dataset\slash code\slash model as part of their submissions via structured templates. This includes details about training, license, limitations, etc. 
        \item The paper should discuss whether and how consent was obtained from people whose asset is used.
        \item At submission time, remember to anonymize your assets (if applicable). You can either create an anonymized URL or include an anonymized zip file.
    \end{itemize}

\item {\bf Crowdsourcing and research with human subjects}
    \item[] Question: For crowdsourcing experiments and research with human subjects, does the paper include the full text of instructions given to participants and screenshots, if applicable, as well as details about compensation (if any)? 
    \item[] Answer: \answerNA{}.
    \item[] Justification: The paper does not involve crowdsourcing experiments or research with human subjects.
    \item[] Guidelines:
    \begin{itemize}
        \item The answer \answerNA{} means that the paper does not involve crowdsourcing nor research with human subjects.
        \item Including this information in the supplemental material is fine, but if the main contribution of the paper involves human subjects, then as much detail as possible should be included in the main paper. 
        \item According to the NeurIPS Code of Ethics, workers involved in data collection, curation, or other labor should be paid at least the minimum wage in the country of the data collector. 
    \end{itemize}

\item {\bf Institutional review board (IRB) approvals or equivalent for research with human subjects}
    \item[] Question: Does the paper describe potential risks incurred by study participants, whether such risks were disclosed to the subjects, and whether Institutional Review Board (IRB) approvals (or an equivalent approval/review based on the requirements of your country or institution) were obtained?
    \item[] Answer: \answerNA{}.
    \item[] Justification: The paper does not involve crowdsourcing, user studies, or other human-subjects research, so IRB approval is not applicable.
    \item[] Guidelines:
    \begin{itemize}
        \item The answer \answerNA{} means that the paper does not involve crowdsourcing nor research with human subjects.
        \item Depending on the country in which research is conducted, IRB approval (or equivalent) may be required for any human subjects research. If you obtained IRB approval, you should clearly state this in the paper. 
        \item We recognize that the procedures for this may vary significantly between institutions and locations, and we expect authors to adhere to the NeurIPS Code of Ethics and the guidelines for their institution. 
        \item For initial submissions, do not include any information that would break anonymity (if applicable), such as the institution conducting the review.
    \end{itemize}

\item {\bf Declaration of LLM usage}
    \item[] Question: Does the paper describe the usage of LLMs if it is an important, original, or non-standard component of the core methods in this research? Note that if the LLM is used only for writing, editing, or formatting purposes and does \emph{not} impact the core methodology, scientific rigor, or originality of the research, declaration is not required.
    \item[] Answer: \answerYes{}.
    \item[] Justification: LLMs are central to the method as compatibility judges, template constructors, and solver-code generators; the Methodology, Experiments, and Appendix prompt section describe these uses.
    \item[] Guidelines:
    \begin{itemize}
        \item The answer \answerNA{} means that the core method development in this research does not involve LLMs as any important, original, or non-standard components.
        \item Please refer to our LLM policy in the NeurIPS handbook for what should or should not be described.
    \end{itemize}

\end{enumerate}

\end{document}